\documentclass[10pt,twocolumn,letterpaper]{article}

\usepackage{cvpr}              % To produce the CAMERA-READY version
\usepackage{algorithm}
\usepackage{algorithmic}
\usepackage{newfloat}
\usepackage{listings}
\usepackage[utf8]{inputenc}
\usepackage{amsmath}
\usepackage{booktabs}
\usepackage{enumitem}  %這個跟itemize衝突
\usepackage{multirow}
\usepackage{tabularray}
\usepackage{bbding}
\usepackage{subcaption}
\usepackage[accsupp]{axessibility} 
\usepackage{xspace}
\usepackage{xcolor}
\usepackage[colorlinks=true,
            linkcolor=black,
            urlcolor=magenta,
            citecolor=cvprblue,
            filecolor=cyan]{hyperref}
  
\definecolor{cvprblue}{rgb}{0.21,0.49,0.74}
\input{custom_environment}
\title{MonoTAKD: Teaching Assistant Knowledge Distillation for Monocular 3D Object Detection}

\author{
Hou-I Liu$^{1,2}$\thanks{This work was done during Hou-I's VISIT program at UW}, 
Christine Wu$^{2}$, 
Jen-Hao Cheng$^{2}$, 
Wenhao Chai$^{2}$, 
Shian-Yun Wang$^{3}$, 
Gaowen Liu$^{4}$,  \\
Hugo Latapie$^{5}$, 
Jhih-Ciang Wu$^{6}$, 
Jenq-Neng Hwang$^{2}$\thanks{Corresponding email: hwang@uw.edu},  
Hong-Han Shuai$^{1}$, 
Wen-Huang Cheng$^{7}$ \\
\vspace{-10pt}
\\
\small
$^{1}$National Yang Ming Chiao Tung University, 
$^{2}$University of Washington, 
$^{3}$University of Southern California,\\ 
\small
$^{4}$Cisco Systems, Inc., 
$^{5}$Taijitu AI, Inc., 
$^{6}$National Taiwan Normal University, 
$^{7}$National Taiwan University
}
\begin{document}
\maketitle

\begin{abstract}
Monocular 3D object detection (Mono3D) holds noteworthy promise for autonomous driving applications owing to the cost-effectiveness and rich visual context of monocular camera sensors. However, depth ambiguity poses a significant challenge, as it requires extracting precise 3D scene geometry from a single image, resulting in suboptimal performance when transferring knowledge from a LiDAR-based teacher model to a camera-based student model. To facilitate effective distillation, we introduce Monocular Teaching Assistant Knowledge Distillation (MonoTAKD), which proposes a camera-based teaching assistant (TA) model to transfer robust 3D visual knowledge to the student model, leveraging the smaller feature representation gap. Additionally, we define 3D spatial cues as residual features that capture the differences between the teacher and the TA models. We then leverage these cues to improve the student model's 3D perception capabilities. Experimental results show that our MonoTAKD achieves state-of-the-art performance on the KITTI3D dataset. Furthermore, we evaluate the performance on nuScenes and KITTI raw datasets to demonstrate the generalization of our model to multi-view 3D and unsupervised data settings. Our code is available at \url{https://github.com/hoiliu-0801/MonoTAKD}.

% Monocular 3D object detection (Mono3D) holds noteworthy promise for autonomous driving applications owing to the cost-effectiveness and rich visual context of monocular camera sensors. However, depth ambiguity poses a significant challenge, as it requires extracting precise 3D scene geometry from a single image, resulting in suboptimal performance when transferring knowledge from a LiDAR-based teacher model to a camera-based student model. To address this issue, we introduce Monocular Teaching Assistant Knowledge Distillation (MonoTAKD) to enhance 3D perception in Mono3D. 
% Our approach presents a robust camera-based teaching assistant model that effectively bridges the representation gap between different modalities for teacher and student models, addressing the challenge of inaccurate depth estimation. 
% By defining 3D spatial cues as residual features that capture the differences between the teacher and the teaching assistant models, we leverage these cues into the student model, improving its 3D perception capabilities. Experimental results show that our MonoTAKD achieves state-of-the-art performance on the KITTI3D dataset. Additionally, we evaluate the performance on nuScenes and KITTI raw datasets to demonstrate the generalization of our model to multi-view 3D and unsupervised data settings. Our code is available at \url{https://github.com/hoiliu-0801/MonoTAKD}.
\end{abstract}
 
\vspace{-10pt}
\setcounter{footnote}{0}
\section{Introduction}
\label{sec:intro}
Monocular 3D object detection (Mono3D)  has garnered significant attention in autonomous driving, driven by the affordability and practicality of monocular camera sensors~\cite{monodetr,monocon,monodtr,monoatt,fusion1}. One widely used approach for improving 3D detection performance is to leverage depth maps as auxiliary supervision. For example, MonoDTR~\cite{monodtr} and MonoPGC~\cite{monopgc} apply an additional depth branch to predict depth maps and integrate them with the visual features using a transformer-based decoder.  
However, the accuracy of depth maps estimated from a single image remains constrained for the camera-based models, primarily due to the absence of stereoscopic data, which is critical to the understanding of 3D scene geometry~\cite{3dunder0,3dunder1}. 

\begin{figure}[t]
\centering
            \includegraphics[width=1
            \columnwidth]{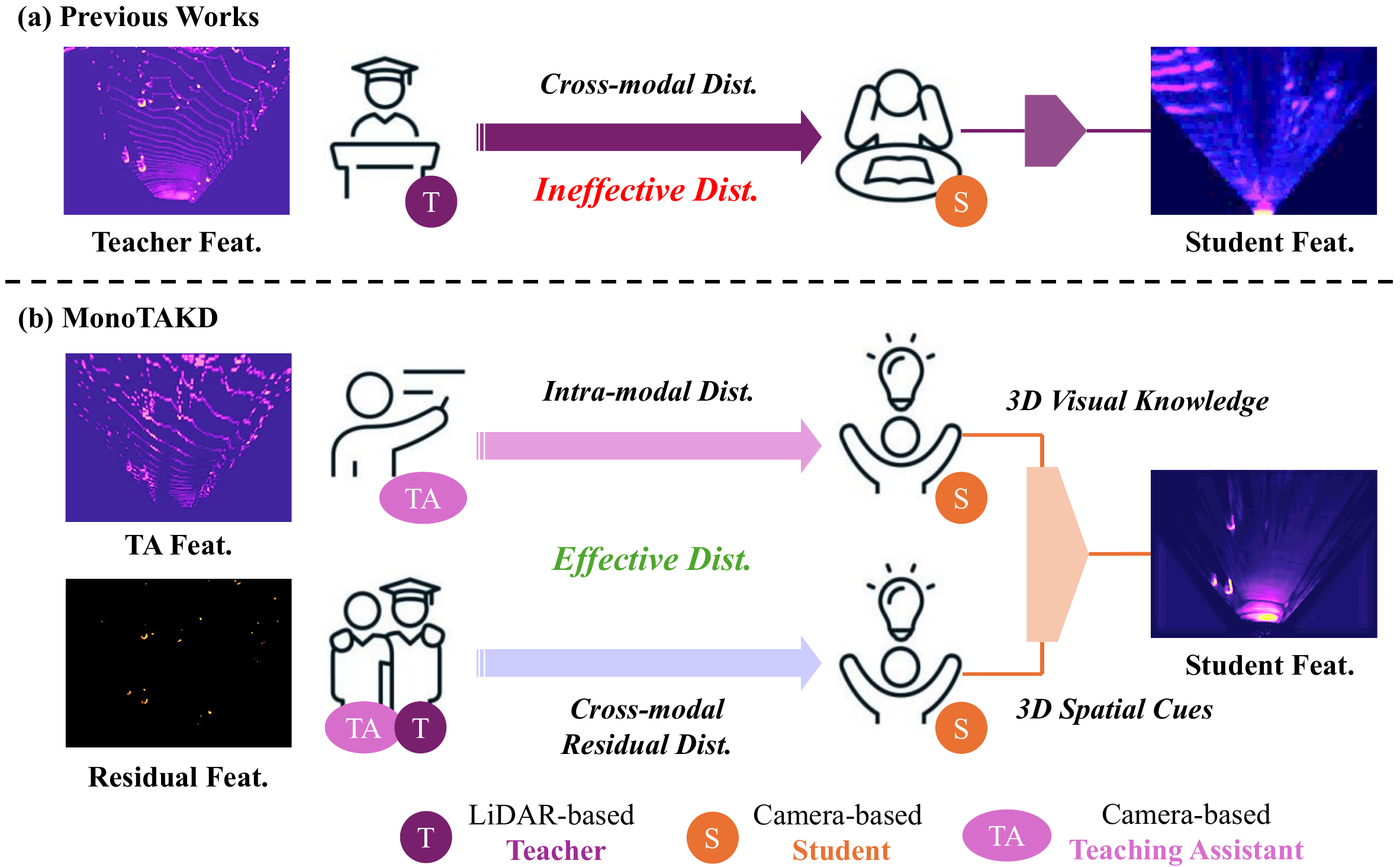}
            \caption{Comparison between previous methods and MonoTAKD (Ours).
            Previous works~\cite{cmkd,monodis} face a significant challenge in the distillation process due to a substantial gap in feature representation.
            Our MonoTAKD incorporates intra-modal and cross-modal residual distillation to enhance learning across this feature representation gap. We visualize the BEV features of each model, along with the residual features derived between the TA and T models, highlighted in orange. Best view with zoom-in and color.}
            \label{fig:intro}
            \vspace{-5pt}
\end{figure}

To enhance the extraction and comprehension of 3D information, a promising alternative to depth-guided approaches is to utilize the cross-modal distillation~\cite{cmkd,unidis}. This approach enables models trained with images to acquire 3D information directly from LiDAR data. For instance, CMKD~\cite{cmkd} and UniDistill~\cite{unidis} employ a LiDAR-based model (teacher) to extract critical 3D information and distill this knowledge into the camera-based model (student) through the bird's eye view (BEV) distillation~\cite{bevdistill}.
Nevertheless, the feature representations from different modalities are inherently distinct, which causes ineffective distillation and restricts the performance of the camera-based student~\cite{2022tig,2023disbev}, as depicted in Fig.~\ref{fig:intro} (a). 

Although recent approaches involve using an adaptation module~\cite{cmkd,2023disbev} or applying regularization~\cite{stxd} to minimize the global distance between cross-modal features, effectively transferring knowledge between such a considerable modality gap continues to be an open challenge.
% To minimize the global distance between cross-modal features, recent approaches involve using an adaptation module~\cite{cmkd,disbev} or applying regularization~\cite{stxd} to map the cross-modal features. 
% Unfortunately, learning an effective alignment remains challenging due to the considerable differences between the two modalities.
Therefore, we introduce a novel Teaching Assistant Knowledge Distillation framework, MonoTAKD. 
% Following the previous cross-modal distillation framework, we use a LiDAR-based model as the teacher and a camera-based model as the student. 
First, we introduce a robust camera-based model as the teaching assistant (TA) model to guide the camera-based student model with 3D visual knowledge, as shown at the top of Fig.~\ref{fig:intro} (b). 
The TA model directly accesses the ground truth (GT) depth map and integrates visual features from the camera-based model to reconstruct optimal BEV features, referred to as 3D visual knowledge. The TA model has two primary functions:
\begin{enumerate*}[label=(\arabic*)]
    \item By providing the optimal BEV features to the student model, it mitigates the distortion brought by inaccurate depth estimation in the camera-based model.
    \item This approach facilitates effective distillation through intra-modal distillation (IMD), as the feature representation gap is smaller within the same modality (camera) compared to cross-modal scenarios (LiDAR and camera)~\cite{add}.
\end{enumerate*}

Specifically, we attempt to make the BEV features of the student model close to those of the TA model\footnote{We conduct knowledge distillation of student and TA models in a unified BEV space to facilitate 3D detection.}. As such, the 3D perception of the student model is enhanced without relying on LiDAR data. It is worth noting that there is no distillation relationship between the teacher and TA models, as visual and LiDAR features serve as independent learning targets. Additionally, we use end-to-end rather than multi-step~\cite{takd} distillation during training. 
% Footnote: Our student and the TA models both project the visual features into the BEV plane.

Even when the student model perfectly replicates the BEV features of the TA model through IMD, a feature representation gap still exists between the replicated BEV features and those of the teacher model. This gap arises because the monocular image inherently lacks certain 3D spatial cues that are exclusive to the LiDAR modality. To this end, we formulate these 3D spatial cues by computing the feature difference in BEV features between the teacher and TA models. Consequently, we define this feature difference as the residual features that are distilled into the student model through the proposed cross-modal residual distillation (CMRD), as illustrated at the bottom of Fig.~\ref{fig:intro} (b). This approach empowers the student model to concentrate on learning the crucial 3D spatial cues instead of being compelled to replicate the complex entirety of BEV features from the LiDAR-based teacher. We believe that learning the key differences (residual features) between the LiDAR and camera model is pivotal for the camera-based student to enhance its 3D perception. 

Camera-based models struggle to extract 3D spatial cues due to their limited understanding of 3D scene geometry. Thus, we propose a spatial alignment module (SAM) to refine the student's BEV features. 
This module enhances the 3D spatial cues by capturing global information and compensating for the spatial shifts, improving the 3D representation of the student's BEV features.
Notably, SAM is an optional module with an acceptable FLOPs overhead. Even without SAM, our MonoTAKD still significantly surpasses state-of-the-art methods by a noticeable margin.

Our contributions are three-fold:
\begin{itemize}
    \item Our MonoTAKD utilizes intra-modal distillation to transfer 3D visual knowledge and cross-modal residual distillation to convey essential 3D spatial cues, both directed to the student model.
    \item We develop a SAM to garner rich global information and to compensate for the spatial shift caused by feature distortion. Also, we design an FFM to expertly integrate features from different modalities, resulting in a more comprehensive 3D representation.
    \item Experimental results show that MonoTAKD achieves state-of-the-art performance on the KITTI3D dataset and validates its generalizability to multi-view and unsupervised settings on the nuScenes and KITTI raw datasets.
\end{itemize}

% MonoTAKD enhances baseline methods by notable margins (+\%) on the nuScenes dataset.
\section{Related Work}
\label{sec:related work}
\def\model {MonoTAKD }

\subsection{LiDAR-based 3D Object Detection}
LiDAR sensors have been widely used in 3D object detection since point clouds can represent precise 3D environmental information~\cite{pvrcnn}. For example, point-based methods~\cite{pointnet,pointnet3} take raw points as input and process the point-wise features with a sizeable multi-layer perceptron. Voxel-based methods~\cite{virconv,vrcnn,second} convert point clouds into voxel grids and extract the voxelized features through 3D sparse convolution layers. Although LiDAR-based 3D object detection methods have proven to be high-performing techniques, LiDAR systems are expensive, making them impractical for autonomous driving.

\subsection{Depth-guided Mono3D}
Monocular 3D object detection methods offer a promising low-cost solution for autonomous driving applications. 
Depth-guided methods~\cite{fusion1,fusion2,caddn,monodetr} leverage depth information to help 3D perception.
Early methods~\cite{fusion1,fusion2} implemented an off-the-shelf depth estimator to predict depth maps, which are then integrated into visual features through a fusion module.
CaDDN~\cite{caddn} predicts depth distribution bins with image features to reconstruct a 3D frustum feature.
However, image features lack the understanding of 3D scene geometry, leading to unreliable depth predictions.

\subsection{Semi-supervised Mono3D}
Semi-supervised Mono3D methods typically incorporate unlabeled data into the training dataset, increasing the quantity of valuable data and thereby enhancing the model's robustness.
LPCG~\cite{lpcg} conducts instant segmentation on images and uses a heuristic algorithm to create 3D pseudo labels. Mix-Teaching~\cite{mixtea} exploits the pre-trained model to generate pseudo labels and pastes them into the image background regions to increase the number of instances in unlabeled data. 
Yet, increasing training data also brings significant training time, and the generated noisy pseudo-labels may hurt the performance.
\begin{figure*}[t]
\centering
    \includegraphics[width=1\textwidth]{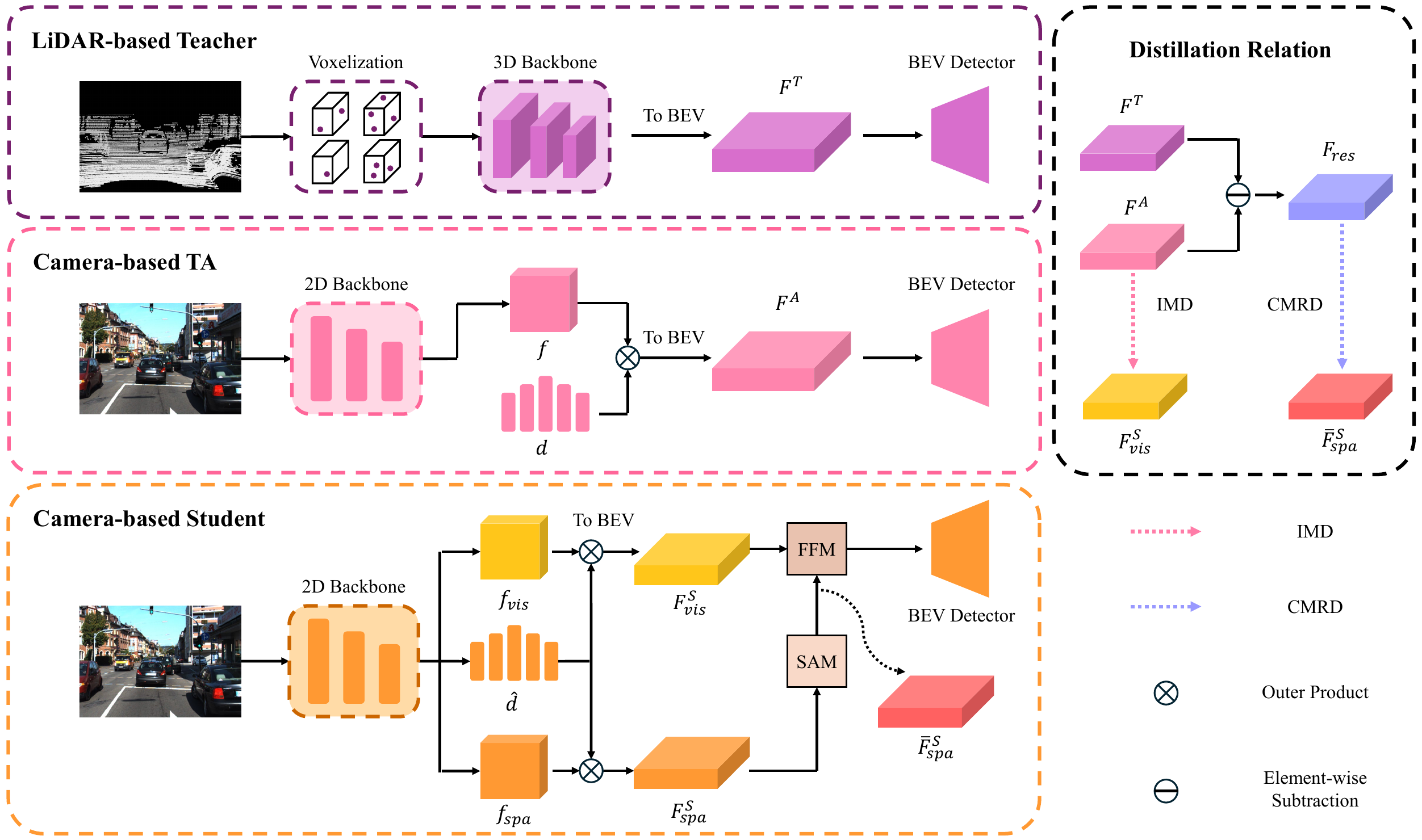}
    \caption{Overall architecture of the~\model. 
    The top, middle, and bottom rows show the architecture of the LiDAR-based teacher, the camera-based teaching assistant (TA), and a camera-based student. 
    We design the intra-modal distillation (IMD) and the cross-modal residual distillation (CMRD) processes to guide the camera-based student. 
    In addition, a spatial alignment module (SAM) and a feature fusion module (FFM) are employed to improve the BEV feature representation.}
\label{fig:overall}  
\end{figure*}

\subsection{Cross-modal Distillation for Mono3D}
Several works~\cite{monodis,cmkd,add,unidis,2023disbev} apply cross-modal distillation to Mono3D.
For example, MonoDistill~\cite{monodis} uses a camera-based model as the teacher model to process the sparse depth map and transfer inferred 3D information to the student model through feature and logit distillation. 
To enhance 3D information, CMKD~\cite{cmkd} and DistillBEV~\cite{2023disbev} employ a LiDAR-based model to extract enriched 3D features from the LiDAR point cloud. Then, they perform the BEV distillation to transfer this knowledge from the LiDAR modality to the camera modality. 

Nevertheless, prior research in this area has predominantly focused on using a single modality teacher, hindering the comprehensive acquisition of 3D knowledge. A camera-based student often struggles to gain this knowledge due to substantial differences in feature representation. Cross-modal feature alignment typically relies on a complex adaptation module, which may be insufficient for effectively bridging this gap~\cite{cmkd,2023disbev}.
In contrast, our~\model employs intra-modal distillation to effectively transfer 3D visual knowledge and introduces cross-modal residual distillation to convey essential 3D spatial cues.

% However, previous knowledge distillation (KD) works have only considered a single modality teacher, potentially restricting the comprehensiveness of 3D knowledge acquisition. For example, the teacher with camera modality may not be able to provide precise 3D information like that of in LiDAR, while the teacher with LiDAR modality may suffer from the large feature representation gap.

% and can better benefit from unlabeled data.

\def\model {MonoTAKD}

\section{Method}
\label{sec:method}

\subsection{Overview}
As shown in Fig.~\ref{fig:overall}, MonoTAKD comprises a pre-trained LiDAR-based model as the teacher $\mathcal{T}$ and two camera-based models, serving as the teaching assistant $\mathcal{A}$ and student $\mathcal{S}$. These three models communicate by propagating BEV features, enabling knowledge distillation (KD) during training to enhance the performance of Mono3D tasks. The core innovation lies in integrating two distillation techniques: intra-modal distillation (IMD), which enriches 3D visual knowledge, and cross-modal residual distillation (CMRD), which provides 3D spatial cues. To further strengthen the BEV feature representation in $\mathcal{S}$, we introduce a spatial alignment module (SAM) to refine its BEV features. Finally, a feature fusion module (FFM) is applied to unify features from both distillation branches of $\mathcal{S}$, ensuring a more cohesive 3D representation.

\subsection{LiDAR-based Teacher}
\label{sec:lidar_teacher}
To exploit the 3D representation and to enhance the 3D perception capabilities of the student model $\mathcal{S}$, we use a pre-trained LiDAR-based model~\cite{second} as the teacher model $\mathcal{T}$ to directly encode 3D information for distillation. As shown in the top block of Fig.~\ref{fig:overall}, $\mathcal{T}$ voxelizes the unordered point cloud data captured from LiDAR into voxel grids $v$, passing through the 3D sparse convolution layers $\psi$ to obtain voxel features. We then employ height compression~\cite{cmkd}, denoted as $\phi$, to embed these voxel features into the BEV space, forming the BEV feature $F^{T} \in\mathbb{R}^{H \times W \times C}$. The process to derive the BEV feature from $\mathcal{T}$ can be formulated as
\begin{equation}
    \label{eq:teacher_bev}
    F^{T} = \phi(\psi(v)),
\end{equation}
% \wu{where $\psi$ represents a series of sparse convolution layers.}

\subsection{Camera-based Teaching Assistant}
% \wu{Cross-modal distillation~\cite{cmkd,monodis} for Mono3D presents challenges due to the intractable heterology of knowledge and the necessity for precise alignment across modalities, making a sophisticated transformation design essential~\cite{bevdistill,stxd}. 
Cross-modal distillation in Mono3D~\cite{cmkd,monodis} is challenging due to the knowledge heterogeneity between modalities, and relying on solely complex adaptation modules~\cite{2023disbev,cmkd} for feature alignment proves insufficient.
To overcome the limitations of direct cross-modal distillation, we introduce a negotiator, \ie, the Teaching Assistant (TA) model, to effectively bridge the teacher and student models, facilitating effective knowledge transfer via IMD. To alleviate knowledge heterogeneity, we select a camera-based model~\cite{caddn} as the TA model $\mathcal{A}$, which shares the same modality as the student model $\mathcal{S}$. We further ensure feature alignment is tractable by jointly projecting the latent representations into BEV space, allowing $\mathcal{S}$ to effectively obtain relevant 3D visual knowledge from $\mathcal{A}$.

As depicted in the middle block of Fig.~\ref{fig:overall}, $\mathcal{A}$ processes a monocular image input by extracting visual features through a 2D backbone (ResNet50), including a $1\times1$ convolutional layer to adjust the channel dimensions. Unlike previous depth-guided methods~\cite{monodetr,monodtr}, $\mathcal{A}$ operates on the ground truth (GT) depth maps $d \in\mathbb{R}^{H \times W \times D}$ by computing the outer product with the resulting visual feature $f$, where $D$ stands for the number of discrete depth bins. This approach effectively injects accurate depth information into the TA model, reducing feature distortion from inaccurate depth estimations and enabling the TA model to achieve performance closer to the optimal levels expected from a camera-based model. Formally, we formulate the BEV feature from $\mathcal{A}$ as
\begin{equation}
    \label{eq:ta_bev}
    F^{A} = \phi(f \otimes d),
\end{equation}
where $d$ represents the GT depth maps, $\otimes$ denotes the outer product, $\phi$ is the projection function similar to Eq.~\ref{eq:teacher_bev} with minor modifications (\eg interpolation~\cite{cmkd}), and $1\times1$ convolution for arranging the desired dimension. 
% The $F^{A}$ represents the composition of ground truth depth maps and visual features from camera-based models. 
Hereafter, we refer to $F^{A}$ as 3D visual knowledge, which serves as one of the learning targets for the student model.
% We hereafter use 3D visual knowledge $F^{A}$ in this work to represent the composition of ground truth depth maps and visual features from camera-based models.

\subsection{Camera-based Student}
The goal in designing the student model is to learn 3D scene geometry from more powerful models. Based on this perspective, we adopt a camera-based model that closely mirrors $\mathcal{A}$. Specifically, we use the same 2D backbone to extract the visual feature and incorporate an off-the-shelf depth estimator to predict depth map $\hat{d}$. After processing the image through the 2D backbone, $\mathcal{S}$ employs two separate branches to generate distinct visual features, denoted as $f_{vis}$ and $f_{spa}$, with different weights.
$f_{vis}$ focuses on learning 3D visual knowledge from $\mathcal{A}$, while $f_{spa}$ is responsible for learning 3D spatial cues from the relationship between $\mathcal{A}$ and $\mathcal{T}$, complementing the visual information. This dual-branch structure optimizes the student's capacity to understand both visual and spatial information crucial for robust 3D perception. Together with the predicted depth map $\hat{d}$, the paired representations go through the projection described in Eq.~\ref{eq:ta_bev}, resulting in BEV features $F^S_{vis}$ and $F^S_{spa}$. 

With the established BEV features ($F^{T}$, $F^{A}$, $F^S_{vis}$, $F^S_{spa}$) from the three models, the following sections detail the distillation, refinement, and fusion processes that drive comprehensive integration of 3D information across modalities.

\noindent \textbf{Intra-modal Distillation.}
Since the camera-based $\mathcal{A}$ and $\mathcal{S}$ share the same modality, their feature representation gap is expected to be smaller than the gap observed in direct cross-modal distillation between $\mathcal{T}$ and $\mathcal{S}$. This similarity in representation makes intra-modal distillation (IMD) a more effective strategy for knowledge transfer. 
In our approach, we leverage the BEV features $F^{A}$ as 3D visual knowledge and distill them into the student's BEV features $F^S_{vis}$. 

The IMD process is optimized by minimizing the mean square error (MSE) between corresponding feature representations, which can be expressed as
\begin{equation}
    \label{eq:IMD}
    \mathcal{L}_{IMD} = \text{MSE}(F^S_{vis},F^{A}).
\end{equation}

\begin{figure}[t]
\centering
            \includegraphics[width=0.7\columnwidth]{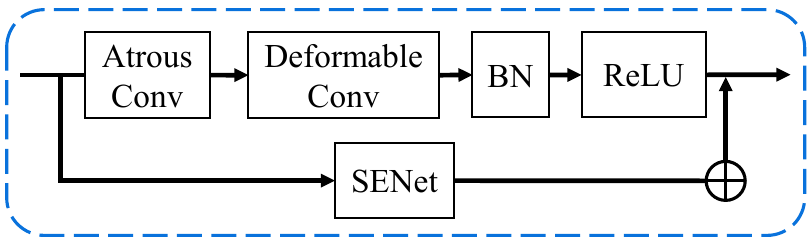}
            \caption{Spatial Alignment Module (SAM). SAM cascades the Atrous and Deformable convolutions to learn the alignment of BEV features. SENet is adopted for channel attention.}
            \label{fig:sam}  
\end{figure}

\noindent \textbf{Spatial Alignment Module.}
To strengthen the student's ability to capture 3D information, we introduce a Spatial Alignment Module (SAM) to refine its BEV features. We denote $\bar{F}^S_{spa}$ as the enhanced BEV features obtained by processing $F^S_{spa}$ through SAM. As illustrated in Fig.~\ref{fig:sam}, SAM captures extensive global information and mitigates spatial feature misalignment through Atrous and Deformable convolutions. More precisely, Atrous Spatial Pyramid Pooling (ASPP)~\cite{aspp} expands the receptive fields using multiscale dilated convolutions, encouraging the model to capture richer 3D information. On the other hand, Deformable convolutions~\cite{dcnv2} adjust spatial offsets to address distortions in BEV features caused by inaccurate depth estimations, thus reducing spatial misalignment in feature representation. Additionally, we incorporate a SENet-like block~\cite{senet} within SAM to recalibrate features adaptively along the channel dimension, enhancing the model's sensitivity to essential 3D spatial cues. This well-designed module allows SAM to enhance feature alignment and effectively integrate valuable spatial information into the student's BEV features. 

\begin{figure}[t]
\centering
            \includegraphics[width=.47\textwidth]{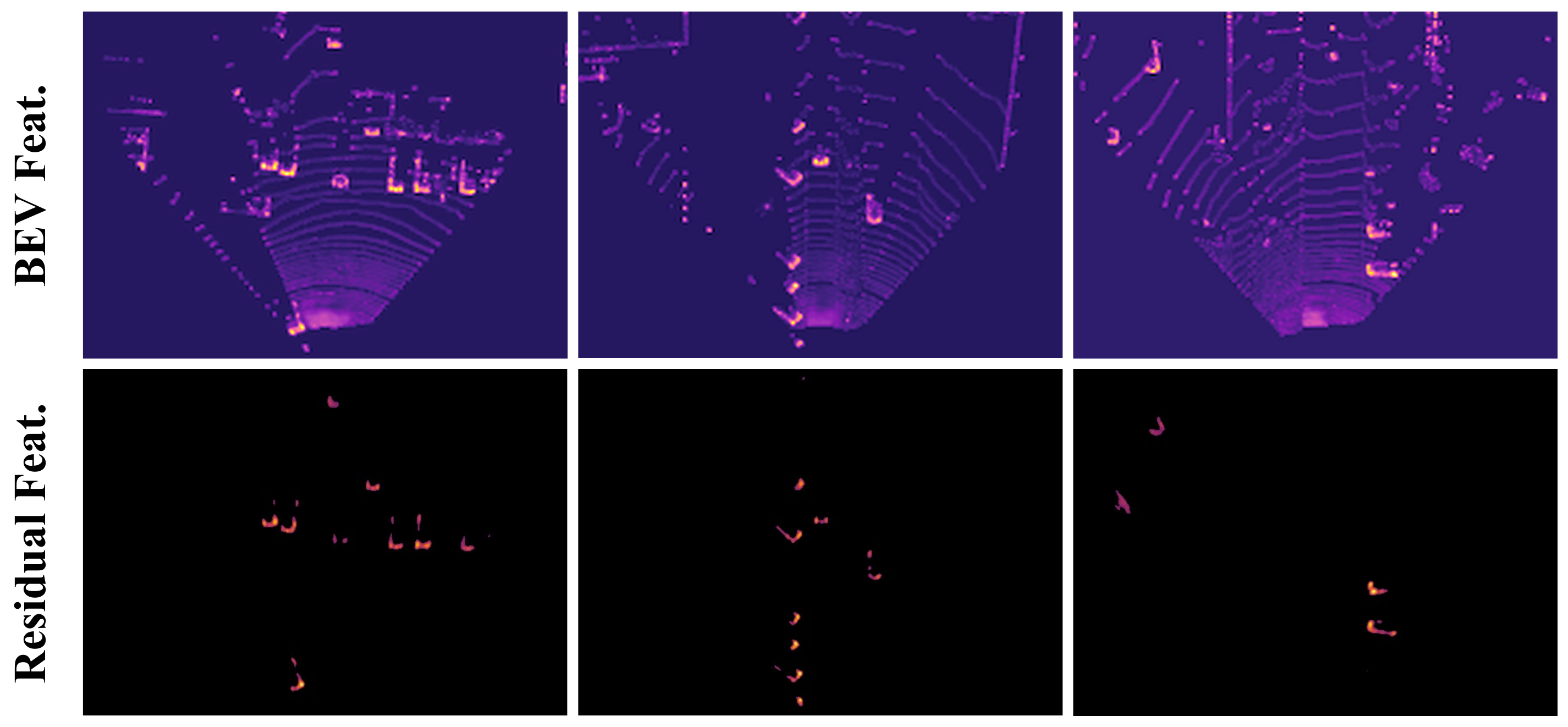}
            \caption{Comparing the BEV features ($F^{T}$) from the teacher model with the residual features ($F_{res}$) reveals that our residuals effectively capture essential 3D spatial cues, emphasizing critical information over less important elements, such as ripples and background noise present in $F^{T}$.}
            \label{fig:vis_res}  
\end{figure}

\noindent \textbf{Cross-modal Residual Distillation.}
While we designate the BEV features of the TA model as the learning target for the student model in Eq.~\ref{eq:IMD}, they still fall short of fully capturing the 3D spatial richness provided by LiDAR. Although incorporating GT depth maps into the TA model narrows such a gap, a discrepancy remains due to the lack of precise 3D spatial cues intrinsic to LiDAR data. Unlike previous cross-modal distillation methods, which require the student model to mimic the raw features from the teacher, our MonoTAKD identifies the missing 3D spatial cues as residual features, providing an additional learning objective. Particularly, the distillation of residual features covers vital spatial information and enhances the 3D perception of the student model. The loss function can be defined as

\begin{equation}
    \label{eq:CMRD}
    \mathcal{L}_{CMRD} = \text{MSE}(\bar{F}^S_{spa},F_{res}),
\end{equation}
where $F_{res} = F^{T} \ominus F^{A}$ represents the residual features. We note that $F_{res}$ is calculated as the difference of BEV features between $\mathcal{T}$ and $\mathcal{A}$ by element-wise subtraction and is binarized by a predefined threshold to suppress background noise and emphasize essential spatial regions (see Fig.~\ref{fig:vis_res}).

\noindent \textbf{Feature Fusion Module.}
To aggregate two BEV features from the student model, \ie, $F^S_{vis}$ and $\bar{F}^S_{spa}$, we apply a Feature Fusion Module (FFM) to integrate their expertise in both visual and spatial 3D information. Specifically, we begin by fusing features $F^S_{vis}$ and $\bar{F}^S_{spa}$ (the enhanced feature by SAM) using element-wise addition. This combined representation is then processed through two convolutional layers. The fusion process can be simplified as
\begin{equation}
    \label{eq:ffm}
        F^{S} =  \text{FFM}(F^S_{vis} \oplus \bar{F}^S_{spa}).
\end{equation}

The resulting fused feature $F^{S}$ is subsequently fed into a BEV detector~\cite{pointpillar} for 3D object detection, generating the prediction for object localization and recognition.

\subsection{Loss Function}
Our proposed MonoTAKD framework is end-to-end trained with an objective function $\mathcal{L}_{total}$ that combines multiple components to effectively guide KD. This total loss function can be expressed as
\begin{equation}
    \label{eq:total_loss}
    \mathcal{L}_{total} = \mathcal{L}_{IMD} + \mathcal{L}_{CMRD} + \mathcal{L}_{logit},
\end{equation}
where the first two terms are introduced in Eq.~\ref{eq:IMD} and Eq.~\ref{eq:CMRD}. The last term in Eq.~\ref{eq:total_loss} is logit distillation for the student model, formulated as
\begin{equation}
    \label{eq:response}
    \mathcal{L}_{logit} = \mathcal{L}_{cls} + \mathcal{L}_{reg},
\end{equation}
where the former term represents the classification loss accomplished by quality focal loss~\cite{qfl}. The latter utilizes the Smooth L1 loss for the bounding box regression for precise localization. Both classification and regression losses are computed over the predictions from the teacher model.
\section{Experiments}
\label{sec:experiment}            

\def\model{MonoTAKD } 
% \definecolor{Gray}{gray}{0.85}
\begin{table*}[ht]
\centering
\scriptsize
\caption{Experimental results on the KITTI \textit{test} set for the Car category. We use \textbf{bold} and \underline{underline} to indicate the best and the second-best results, respectively. \dag~denotes the KD-based methods.}
\vspace{-5pt}
\resizebox{0.9\linewidth}{!}{
\begin{tabular}{l|c|c|ccc|ccc}
% \small
\hline
\multirow{2}{*}{Method}  & \multirow{2}{*}{Venue} & \multirow{2}{*}{Extra Data} & \multicolumn{3}{c|}{$AP_{3D}$} &  \multicolumn{3}{c}{$AP_{BEV}$}     \\
% \cline{4-6} \cline{7-9}
                         &         &            & Easy  & Mod.  & Hard  & Easy   & Mod.  & Hard  \\
\hline
DDMP-3D~\cite{ddmp}       & CVPR 21 & \multirow{2}{*}{Pre-trained Depth}      & 19.71 & 12.78 & 9.80  & 28.08  & 17.89 & 13.44 \\
DD3D~\cite{dd3d}          & ICCV 21 &            & 23.22 & 16.34 & 14.20 & 30.98  & 22.56 & 20.03 \\
\hline
Kinematic3D~\cite{kin3d}  & ECCV 20 & \multirow{2}{*}{Temporal}   & 19.07 & 12.72 & 9.17  & 26.69  & 17.52 & 13.10 \\
Dfm~\cite{kin3d}          & ECCV 22   & & 22.94 & 16.82 & 14.65  & 31.71 & 22.89 & 19.97 \\
% \hline
% AutoShape~\cite{autoshape}& ICCV 21 & \multirow{2}{*}{CAD}        & 22.47 & 14.17 & 11.36 & 30.66  & 20.08 & 15.95 \\
% DCD~\cite{dcd}            & ECCV 22  &            & 23.81 & 15.90 & 13.21 & 32.55  & 21.50 & 18.25 \\
\hline
CaDDN~\cite{caddn}        & CVPR 21 & \multirow{5}{*}{LiDAR Auxiliary}      & 19.17 & 13.41 & 11.46 & 27.94  & 18.91 & 17.19 \\
MonoDTR~\cite{monodtr}    & CVPR 22 &            & 21.99 & 15.39 & 12.73 & 28.59  & 20.38 & 17.14 \\
MonoNerd~\cite{mononerd}  & ICCV 23 &            & 22.75 & 17.13 & 15.63 & 31.13  & 23.46 & 20.97 \\   
MonoPGC~\cite{monopgc}    & ICRA 23 &            & 24.68 & 17.17 & 14.14 & 32.50  & 23.14 & 20.30 \\
OccupancyM3D~\cite{occm3d} & CVPR 24 &      & 25.55 & 17.02 & 14.79 & 35.38 & 24.18 & 21.37 \\
% DD3DV2~\cite{monopgc}    & ICRA 23 &            & 26.36 & 17.61 & 15.32 & 35.70  & 24.67 & 21.73 \\
\hline
% GUPNet~\cite{gupnet}      & ICCV 21 & \multirow{11}{*}{None}       & 20.11 & 14.20 & 11.77 & 30.29  & 21.19 & 18.20 \\
% DEVIANT~\cite{deviant}    & ECCV 22 &            & 21.88 & 14.46 & 11.89 & 29.65  & 20.44 & 17.43 \\
% MonoCon~\cite{monocon}    & AAAI 22 &   \multirow{11}{*}{None}    & 22.50 & 16.46 & 13.95 & 31.12  & 22.10 & 19.00 \\
MonoDistill\dag~\cite{monodis} & ICLR 22 &  \multirow{8}{*}{None}    & 22.97 & 16.03 & 13.60 & 31.87  & 22.59 & 19.72 \\
Cube R-CNN~\cite{omni-cube} & CVPR 23 &     & 23.59 & 15.01 & 12.56 & 31.70  & 21.20 & 18.43 \\
MonoUNI~\cite{monouni} & NeurIPS 23 &          & 24.75 & 16.73  & 13.49 & 33.28 & 23.05 & 19.39  \\
MonoATT~\cite{monoatt}    & CVPR 23 &       & 24.72 & \underline{17.37} & 15.00 & \underline{36.87}& \underline{24.42} & \underline{21.88} \\
% MonoDDE~\cite{monodde}    & CVPR 22 &       & 24.93 & 17.14 & 15.10 & 33.58  & 23.46 & 20.37 \\
MonoDETR~\cite{monodetr}  & ICCV 23 &       & 25.00 & 16.47 & 13.58 & 33.60  & 22.11 & 18.60 \\
CMKD\dag~\cite{cmkd}           & ECCV 22 &      & 25.09 & 16.99 & \underline{15.30} & 33.69  & 23.10 & 20.67 \\
ADD\dag~\cite{add}             & AAAI 23 &      & \underline{25.61} & 16.81 & 13.79 & 35.20  & 23.58 & 20.08 \\
MonoCD~\cite{MonoCD}       & CVPR 24 &      & 25.53 & 16.59 & 14.53 & 33.41 & 22.81 & 19.57 \\
% MonoMAE~\cite{MonoCD}       & NeurIPS 24 &      & 25.60 & 18.84 & 16.78 & 34.15 & 24.93 & 21.76 \\

\hline
\textbf{\model} & -       & None       & \textbf{27.91} & \textbf{19.43} & \textbf{16.51} & \textbf{38.75} & \textbf{27.76} & \textbf{24.14} \\ 
\hline
\end{tabular}
}
\label{tab:main}
\vspace{-6pt}
\end{table*}

\subsection{Datasets and Evaluation Metrics}
\textbf{KITTI3D}.
% \textbf{Benchmark: KITTI 3D}.
The KITTI 3D detection dataset~\cite{kitti} is a widely used benchmark in the field of autonomous driving. It consists of 7,481 training images and 7,518 testing images with synchronized LiDAR point clouds.% and stereo RGB images.
Following the data split provided by~\cite{val}, the training images are further split into two groups: 3,712 images and 3,769 images as the train and val sets, respectively. 
% \noindent \textbf{Evaluation Metrics.}
% The results are presented on the KITTI test, val, and unlabeled set for the Car category.
We report the 3D detection performance $AP_{3D}$ and the BEV detection performance $AP_{BEV}$ with 40 recall positions~\cite{r40}.
Our detection results include three difficulty levels: easy, moderate, and hard.
% The IoU thresholds for the Car, Pedestrian, and Cyclist categories are 0.7, 0.5, and 0.5, respectively.
% Additionally, we utilize the nuScenes Detection Score (NDS) and the mean Average Precision (mAP) as the primary evaluation metrics for the nuScenes dataset.

% \vspace{6pt}
\noindent \textbf{nuScenes}.
The nuScenes dataset~\cite{nuscene} includes 1,000 multimodal video sequences, 700 for training, 150 for validation, and 150 for testing. 
It offers synchronized sensor data streams collected with 6 cameras and a 32-beam LiDAR sampled at 20Hz, covering the 360-degree field of view. 
We utilize nuScenes Detection Score (NDS) and the mean Average Precision (mAP) as the primary evaluation metrics. 
Due to space limits, other dataset descriptions and implementation details are provided in the supplementary material.

% \subsection{Implementation details} 
% For the KITTI dataset, we use a pre-trained Second~\cite{second} as the LiDAR-based teacher, while both the camera-based teaching assistant and camera-based student are derived from CaDDN~\cite{caddn} and using ResNet50 as the backbone.
% In addition, we use PointPillar~\cite{pointpillar} as the BEV detector. 
% Initially, we trained the teaching assistant model for 30 epochs. Then, a pre-trained teacher model and a frozen teaching assistant model are used to train the student model for another 60 epochs.
% Training is performed with an NVIDIA Titan XP GPU in an end-to-end manner. We set the batch size to 2, and the learning rate is $2e^{-4}$ with the one-cycle learning rate strategy.
% As for the discrete depth bins $D$, we set $D$ to 120, and the minimum and maximum depths are set to 2.0 and 46.8 meters, respectively. 
% Besides, we use $\lambda_{1}$=0.7 and $\lambda_{2}$=0.3 as coefficients for intra-modal distillation and residual feature distillation, respectively.

% For the nuScenes dataset, we adopt a pre-trained CenterPoint~\cite{centerpoint} as the LiDAR-based teacher and use BEVDepth~\cite{bevdepth} as the teaching assistant and student models. 
% Due to a higher resolution and a larger model size, we set the batch size to 8 and trained the models with eight NVIDIA V100 GPUs.
% We set the learning rate of $2e^{-4}$ with a multi-step learning rate decay schedule and a decay rate of 0.1 and train the model for 25 epochs.

\begin{table}
\centering
\caption{Experimental results on the nuScenes \textit{val} set. We use \textbf{bold} to indicate the best results in each setting.}
\vspace{-5pt}
\resizebox{0.99\linewidth}{!}{
\begin{tabular}{l|c|c|cc} 
\hline
Method        & Modality & Backbone & NDS$\uparrow$   & mAP$\uparrow$    \\ 
\hline
PETRv2~\cite{petrv2}     & C        & R50      & 0.456 & 0.349  \\
P2D~\cite{p2d}           & C        & R50      & 0.474 & 0.360  \\
% DETR3D~\cite{detr3d}     & C        & R50      & 0.373 & 0.302  \\
% DETR3D~\cite{detr3d}     & C        & R101     & 0.425 & 0.346  \\
PGD~\cite{PGD}           & C        & R101     & 0.428 & 0.369  \\
MonoDETR~\cite{monodetr}       & C        & R101     & 0.526 & 0.423  \\ 
% Sparse4D                 & C        & R101      & 0.436 & 0.541   \\
\hline
BEVFormer~\cite{bevformer}     & C        & R50      & 0.423 & 0.352  \\
+BEVDistill~\cite{bevdistill}    & $L \rightarrow C$      & R50      & 0.457 & 0.386  \\
% +MonoDistill~\cite{bevformer}    & $L \rightarrow C$      & R50      & 0.429 & 0.364  \\
+DistillBEV~\cite{2023disbev}        & $L \rightarrow C$      & R50      & 0.476 & 0.367  \\

+STXD~\cite{stxd}    & $L \rightarrow C$      & R50      & 0.481 & 0.374  \\
+TAKD (Ours)    & $L \rightarrow C$      & R50      & \textbf{0.490} & \textbf{0.392}  \\ 
\hline
BEVFormer     & C                      & R101     & 0.445 & 0.374  \\
+BEVDistill   & $L \rightarrow C$      & R101     & 0.468 & 0.389  \\
+DistillBEV  & $L \rightarrow C$      & R101     & 0.545 & 0.446  \\
+STXD         & $L \rightarrow C$      & R101     & 0.543 & 0.440   \\
+TAKD (Ours)        & $L \rightarrow C$      & R101     & \textbf{0.558} & \textbf{0.451}  \\ 
\hline
BEVDepth~\cite{bevdepth}      & C                      & R50      & 0.440 & 0.317  \\
% +BEVDistill   & $L \rightarrow C$      & R50     & 0.452 & 0.330  \\
+STXD         & $L \rightarrow C$      & R50     & 0.483 & 0.371   \\
+DistillBEV   & $L \rightarrow C$      & R50      & 0.510  & 0.403  \\
+LabelDistill~\cite{labeldis} & $L \rightarrow C$      & R50      & 0.528 & 0.419  \\
+TAKD (Ours)        & $L \rightarrow C$      & R50      & \textbf{0.537} & \textbf{0.430}   \\ 
\hline
BEVDepth      & C        & R101     & 0.535 & 0.412  \\
+DistillBEV  & $L \rightarrow C$      & R101     & 0.547 & 0.450   \\
+LabelDistill & $L \rightarrow C$      & R101     & 0.553 & 0.451  \\
+TAKD (Ours)         & $L \rightarrow C$      & R101     & \textbf{0.564} & \textbf{0.466}  \\
\hline
\end{tabular}
}
\vspace{-12pt}
\label{tab:nus}
\end{table}

% \vspace{-2em}

\begin{table*}[ht]
\centering
\caption{Compared the effect of distillation losses using various teacher, TA, and student models. * indicates the insertion of the ground truth depth. We use \textbf{bold} and \underline{underline} to indicate the best and the second-best results, respectively.}
\scriptsize
\resizebox{0.9\linewidth}{!}{
\begin{tabular}{l|c|c|ccc|ccc}
\hline
\multicolumn{3}{c|}{Model Types}          & \multicolumn{3}{c|}{$AP_{3D}$} & \multicolumn{3}{c}{$AP_{BEV}$} \\ 
\hline
Teacher model & TA model  & Student model & Easy  & Mod.  & Hard    & Easy & Mod. & Hard   \\ 
\hline
None          & None       & CaDDN         & 23.57 & 16.31 & 13.84     & 30.28 & 21.53 & 18.90  \\ 
CenterPoint~\cite{centerpoint}   & CaDDN*     & CaDDN         & 27.06 & 19.38 & 17.50 & 35.66 & 26.03 & 23.09  \\
PointPillar~\cite{pointpillar}   & CaDDN*     & CaDDN         & 31.28 & 20.80 & 17.58 & \underline{42.21} & \underline{28.48} & \underline{25.73} \\
Second~\cite{second}             & CaDDN*     & CaDDN        & \textbf{34.36} & \textbf{22.61} & \textbf{19.88} & \textbf{42.86} & \textbf{29.41} & \textbf{26.47} \\ 
CenterPoint   & MonoDETR*     & CaDDN         & 26.56 & 18.84 & 16.46    & 34.62 & 25.85 & 22.29   \\
PointPillar   & MonoDETR*     & CaDDN         & 28.11 & 20.00 & 17.24    & 40.21 & 27.10 & 24.65  \\
Second        & MonoDETR*     & CaDDN         & 30.74 & 20.35 & 17.69    & 41.75 & 28.95 & 24.80   \\ 
\hline
None          & None         & MonoDETR        & 28.84 & 20.61 & 16.38   & 37.86 & 26.95 & 22.80   \\
CenterPoint   & MonoDETR*    & MonoDETR        & 30.78 & 21.17 & 18.41   & 39.73 & 27.22 & 24.69   \\
PointPillar   & MonoDETR*    & MonoDETR         & 31.25 & 21.47 & \underline{18.57}   & 38.68 & 27.16 & 24.83  \\
Second        & MonoDETR*    & MonoDETR        & \underline{33.18} & \underline{21.97} & 18.55   & 41.98 & 28.43 & 25.24   \\ 
\hline
\end{tabular}
}
\vspace{-12pt}
\label{tab:diff_TA_teacher}
\end{table*}

\subsection{Main Results}
% \noindent \textbf{KITTI3D.}
\noindent \textbf{Results on KITTI3D.}
We demonstrate the effectiveness of our~\model on the KITTI test set for the Car category in Table~\ref{tab:main}.
First,~\model outperforms the second-best result by +2.30, +2.06, and +1.21 in $AP_{3D}$, by +1.88, +3.34, +2.26 in $AP_{BEV}$ under easy, moderate, and hard difficulty levels, respectively.
When compared to CMKD, the top-performing KD-based method, MonoTAKD, consistently surpasses CMKD, achieving +2.82, +2.44, +1.21 improvements in $AP_{3D}$ across all difficulty levels.
Moreover, we observe a significant performance gain over all proceeding depth-guided methods. For instance, with respect to MonoDETR, our model achieves a notable increase of +2.93 (21.58\%) in $AP_{3D}$ under the hard level.

Our successful performance lies in incorporating intra-modal distillation (IMD) and cross-modal residual distillation (CMRD), which provide the student model with rich 3D visual knowledge and crucial 3D spatial cues. This combined approach enhances both the semantic information and understanding of 3D scene geometry for the student model. Note that the results for other classes (Pedestrian and Cyclist) are provided in the supplementary material.

\noindent \textbf{Results on nuScenes.}
We further evaluate MonoTAKD on the nuScenes val set, as shown in Table~\ref{tab:nus}. For consistent and effective distillation, we use CenterPoint as the LiDAR-based teacher and employ CaDDN for both the TA and student models, as validated in Table~\ref{tab:diff_TA_teacher}. 

Our results show significant improvements in NDS and mAP for both BEVFormer and BEVDepth. 
This reveals the potential to scale our feature distillation losses from monocular to multi-view camera applications, which showcases the generalizability of our approach.
Additionally, we observe higher performance using BEVDepth as the student model due to its similarity to CenterPoint's dense prediction head, enhancing the effectiveness of logit distillation.

\subsection{Ablation Study}
For easier comparison with other methods, we conduct ablation studies on the KITTI val set for the Car category, using ResNet50 as the backbone for the TA and student models.

\vspace{6pt}
% In this section, we demonstrate the effectiveness of each component of the proposed~\model on the KITTI val set for the Car category, using $AP_{3D}$ and $AP_{BEV}$ metrics for comparisons and discussion. 
\noindent \textbf{Generalization Study with Different Teacher and Teaching Assistant Models.}
We employ CaDDN and MonoDETR without distillation techniques as the baseline models in this experiment.
The findings in Table~\ref{tab:diff_TA_teacher} suggest utilizing the same model for the TA and student models. As such, the implementation is simpler and more generalizable than creating a custom TA model for each detector.
In addition, we observe that distilling knowledge across heterogeneous architectures (transformer to CNN) may be ineffective due to substantial differences in feature encoding.
Our study demonstrates that MonoTAKD consistently boosts performance under the guidance of various LiDAR-based teacher and camera-based TA models. 
Ultimately, Second is selected as our LiDAR-based teacher, with CaDDN serving as both the TA and student models.
% \vspace{-1pt}
% the two main features distilled to the student.
% \vspace{6pt}

\noindent \textbf{Effectiveness of Different Feature Distillation.}
We present the effect of each feature distillation loss on MonoTAKD's performance in terms of $AP_{3D}$, as shown in Table~\ref{tab:featdis}.
Setting 1 represents the baseline performance of the student without any distillation loss guidance.
Cross-modal distillation (CMD) refers to the student learning directly from the LiDAR BEV features~\cite{cmkd}.
Settings 2 and 3 investigate the effectiveness of distillation between IMD and CMD. The student model performs better with IMD because the distillation is across a narrower feature representation gap than CMD.
Settings 3 and 4 show that CMRD slightly outperforms CMD, although it does not provide the stable and robust 3D visual knowledge that IMD offers. 

To further highlight the effect of residual features, we incorporate IMD to support the basic 3D perception and compare the results between settings 5 and 6. Our analysis reveals that focusing on learning 3D spatial cues (residual features) within the LiDAR BEV feature is more effective than exhaustively replicating all 3D information from the LiDAR modality. Specifically, the student model can prioritize accurately discerning the shape and position of foreground objects while minimizing attention to background noise, which could otherwise hinder its understanding of 3D scene geometry.
As a result, setting 6 (MonoTAKD) outperforms baseline, setting 1, in $AP_{3D}$ by +10.01, +6.45, and +6.44 for three difficulty levels, respectively.
In our settings, the FFM is excluded from settings 1-4 since the BEV feature fusion is not required.
\begin{table}[!t]
\centering
% \scriptsize
 % on the KITTI \textit{val} set
\caption{Effectiveness of different feature distillation losses.}
\vspace{-5pt}
\resizebox{.9\linewidth}{!}{
\begin{tabular}{c|ccc|ccc} 
\hline
\multirow{2}{*}{Settings} & \multicolumn{3}{c|}{Loss} & \multicolumn{3}{c}{$AP_{3D}$} \\ 
\cline{2-4}
                          & IMD & CMD & CMRD                   & Easy & Mod. & Hard       \\ 
\hline
1                         &     &     &                       & 24.35 & 16.16 & 13.44  \\
2                         & \checkmark   &     &              & 31.11  &20.24  &16.91   \\
3                         &     & \checkmark   &              & 28.22  &18.29  &15.10   \\
4                         &     &    & \checkmark             & 29.68  &19.57  &16.22  \\
5                         & \checkmark   & \checkmark   &     & 30.73 & 19.88 & 16.43  \\
6                         & \checkmark   &     & \checkmark   &\textbf{34.36} & \textbf{22.61} & \textbf{19.88}   \\
\hline
\end{tabular}
}
\label{tab:featdis}
\vspace{-6pt}
\end{table}

% \begin{table*}[!t]
% \centering
% \caption{Effectiveness of components in feature distillation on the KITTI \textit{val} set.}
% % \resizebox{0.85\linewidth}{!}{
% \begin{tabular}{c|ccc|ccc|ccc} 
% \hline
% \multirow{2}{*}{Settings} & \multicolumn{3}{c|}{Loss} & \multicolumn{3}{c|}{$AP_{3D}$}     & \multicolumn{3}{c}{$AP_{BEV}$}  \\ 
% \cline{2-4}
%                           & IMD & CMD & CMRD                   & Easy & Mod. & Hard    & Easy & Mod. & Hard     \\ 
% \hline
% 1                         &     &     &                       & 24.35 & 16.16 & 13.44 & 31.78 &  21.61  & 18.80  \\
% 2                         & \checkmark   &     &              & 31.11  &20.24  &16.91 & 39.77 & 27.03 & 23.22  \\
% 3                         &     & \checkmark   &              & 28.22  &18.29  &15.10 & 36.48  & 24.87 & 21.30  \\
% 4                         & \checkmark   & \checkmark   &     & 30.73 & 19.88 & 16.43 & 39.24&  26.31 & 22.52  \\
% 5                         & \checkmark   &     & \checkmark   &\textbf{34.36} & \textbf{22.61} & \textbf{19.88} & \textbf{42.86} & \textbf{29.41} & \textbf{26.47}  \\
% \hline
% \end{tabular}
% % }
% \label{tab:featdis}
% \end{table*}

\begin{figure}[t]
    \centering
    \hspace{-0.5cm}
    \includegraphics[width=.45\textwidth]{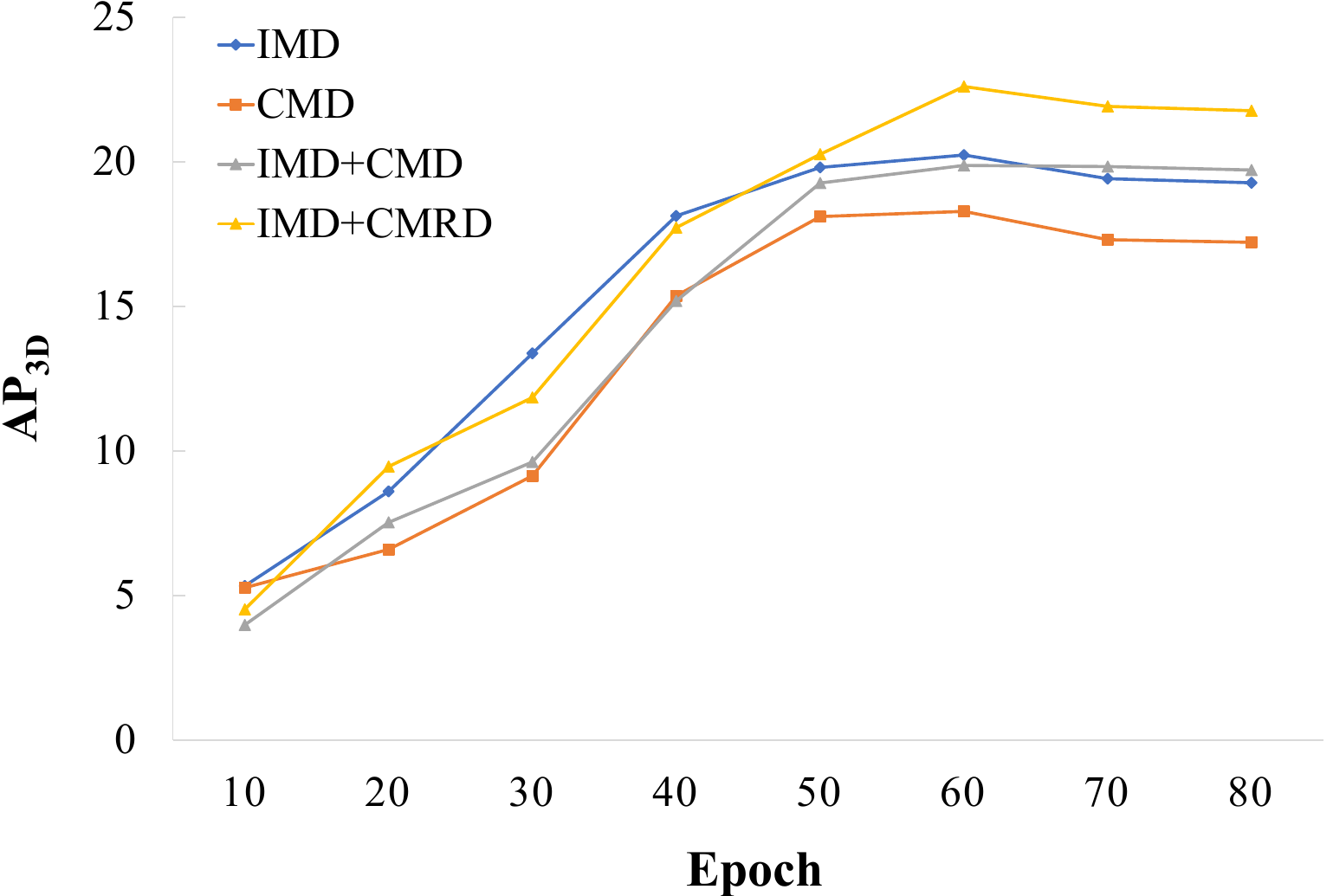}
    \caption{Convergence curves of different feature distillation on the KITTI \textit{val} set. The x-axis shows the number of epochs, and the y-axis denotes $AP_{3D}$ for the Car category at the moderate level. 
    % Note the performance at 60 epochs is highlighted, representing the peak $AP_{3D}$ across all settings.
    }
\label{fig:eff_tab}  
\vspace{-5pt}
\end{figure}

Additionally, we report training curves for each feature distillation loss at the moderate level, as depicted in Fig.~\ref{fig:eff_tab}.
When comparing the training curves between IMD and CMD, CMD consistently shows lower $AP_{3D}$ values. This underscores the superior effectiveness of IMD over CMD, attributed to the narrower feature representation gap within the same modality.

Finally, our IMD+CMRD (MonoTAKD) remarkably outperforms IMD+CMD throughout the training process. This observation indicates that the student model greatly benefits from learning residual features rather than learning from the entirety of LiDAR BEV features, thereby enhancing the effectiveness of CMD.
Notably, each feature distillation loss converges within approximately 60 epochs, demonstrating that the proposed feature distillation losses and modules do not extend training time. Furthermore, all experiments can be conducted on a consumer-grade GPU with only 12GB VRAM, indicating that our method imposes minimal impact on training costs.

\begin{table}[t]
\centering
% on the KITTI \textit{val} set
\caption{Ablation of the components of MonoTAKD. DConv. indicates the deformable convolution.}
% \vspace{5pt}
\scriptsize
\resizebox{\linewidth}{!}{
\begin{tabular}{cccc|ccc} 
\hline
\multicolumn{4}{c|}{Components} & \multicolumn{3}{c}{$AP_{3D}$}  \\ 
\hline
ASPP  & DConv. & SENet & FFM      & Easy  & Mod.  & Hard        \\ 
\hline
&  &  &  &  31.36 & 20.80  & 16.53        \\
\checkmark &        &            &  & 32.28 & 21.45 & 17.18          \\
      & \checkmark  &            &  & 32.46 & 21.51 & 17.99          \\
\checkmark & \checkmark  &       &  & 33.19 & 21.81 & 18.23    \\
\checkmark & \checkmark  & \checkmark &  & 33.26 & 21.47 & 18.61      \\
\checkmark & \checkmark  & \checkmark & \checkmark  & \textbf{34.36} & \textbf{22.61} & \textbf{19.88}      \\
\hline
\end{tabular}
}
\vspace{-5pt}
\label{tab:sam}
\end{table}

\begin{table}[t]
\centering
\caption{Efficiency analysis of MonoTAKD.}
% \vspace{-5pt}
\resizebox{\linewidth}{!}{
\begin{tabular}{l|c|c|ccc} 
\hline
\multirow{2}{*}{Model} & \multirow{2}{*}{\begin{tabular}[c]{@{}c@{}}Parameters\\(M)\end{tabular}} & \multirow{2}{*}{\begin{tabular}[c]{@{}c@{}}FLOPs\\ (G)\end{tabular}} & \multicolumn{3}{c}{$AP_{3D}$} \\ 
\cline{4-6}
 &  &  & Easy & Moderate & Hard \\ 
\hline
MonoNeRD~\cite{mononerd} & 83.0 & 356.57 & 20.64 & 15.44 & 13.99 \\ 
DD3Dv2~\cite{dd3dv2} & 80.3 & 163.00 & 26.23 & 21.21 & 18.83 \\ 
MonoDETR~\cite{monodetr} & 47.4 & 57.51 & 28.84 & 20.61 & 16.38 \\ 
% OccupancyM3D~\cite{occm3d} & 48.3 & 327.52 & 26.87 & 19.96 & 17.15 \\ 
CMKD~\cite{cmkd} & \textbf{45.1} & \textbf{41.32} & 23.53 & 16.33 & 14.44 \\ 
\hline
MonoTAKD-Lite & \textbf{45.1} & \textbf{41.32} & 31.36 & 20.80 & 16.53 \\ 
MonoTAKD & 47.8 & 44.90 & \textbf{34.36} & \textbf{22.61} & \textbf{19.88} \\ 
\hline
\end{tabular}
}
\vspace{-5pt}
\label{tab:eff}
\end{table}

% mononerd, 0.26 fps

% \begin{table}[t]
% \centering
% % \scriptsize
% % on the KITTI \textit{val} set.
% \caption{Efficiency analysis of MonoTAKD.}
% \resizebox{\linewidth}{!}{
% \begin{tabular}{l|c|c|c|ccc} 
% \hline
% \multirow{2}{*}{Model} & \multirow{2}{*}{\begin{tabular}[c]{@{}c@{}}Param.\\(M)\end{tabular}} & \multirow{2}{*}{\begin{tabular}[c]{@{}c@{}}FLOPs\\ (G)\end{tabular}} & \multirow{2}{*}{FPS} & \multicolumn{3}{c}{$AP_{3D}$}  \\ 
% \cline{5-7} & & &      & Easy  & Mod.  & Hard      \\ 
% \hline
% DD3D~\cite{dd3d} &  80.32  & 163.0 &  16.2 & 26.23 &  21.21 & 18.83   \\
% MonoDETR~\cite{monodetr}  & 127.8  & 656.4 & 3.6  & 28.84 & 20.61 & 16.38   \\
% OccupancyM3D~\cite{occm3d} & 28.3   & 327.52 & 5.3  & 26.87 & 19.96 & 17.15   \\
% % CMKD                   & 45.1   & 368.7 
% CMKD~\cite{cmkd}      & \textbf{45.1}   & \textbf{41.32} & \textbf{12.5} & 23.53 & 16.33 & 14.44   \\
% \hline
% MonoTAKD-Lite           & \textbf{45.1}   & \textbf{41.32} & \textbf{12.5} & 30.36 & 19.80 & 16.53  \\
% MonoTAKD            & 49.8   & 411.9 & 11.9 & \textbf{34.36} & \textbf{22.61} & \textbf{19.88} \\
% \hline
% \end{tabular}
% }
% \vspace{-6pt}
% \label{tab:eff}
% \end{table}

% \input{tab/tab_backbone}

\noindent \textbf{Effectiveness of the Proposed Modules.}
Table~\ref{tab:sam} illustrates the performance boost brought by each component of SAM and the FFM. The baseline is the~\model with IMD and CMRD but without SAM and FFM.
In our experiment, FFM is designed to augment BEV feature fusion across different branches of the student model, whereas configurations without it default to element-wise addition.

With the combination of all components,~\model improves +3.00, +1.81, and +3.35 in $AP_{3D}$ across three difficulty levels, highlighting the effectiveness of SAM in learning spatial shifts in both near and far regions, and FFM in integrating BEV features within the student model.

\noindent \textbf{Efficiency Analysis.} 
% \label{tab:eff_an}
To further evaluate the practicality of the model at the application level, we tabulate the FLOPs and parameters of state-of-the-art Mono3D methods in Table~\ref{tab:eff}. 
By leveraging the KD-based framework, CMKD reaches a substantial reduction in parameters and FLOPs compared to depth-guided methods~\cite{dd3dv2,monodetr}.
However, distilling knowledge across the large feature gap between teacher and student models results in a notable AP drop. 

To ensure a fair comparison with CMKD, we remove both SAM and FFM modules from our student model, creating a lightweight version, MonoTAKD-lite, which matches the baseline in Table~\ref{tab:sam} and aligns with the CMKD (CaDDN) architecture. MonoTAKD-Lite effectively mitigates the performance drop while remaining a compact student model. 
Notably, even without the additional modules, our MonoTAKD-Lite demonstrates a clear performance gain over state-of-the-art approaches, offering higher $AP_{3D}$ with reduced FLOPs and fewer parameters. 
% The analysis of the backbone choices can be found in the supplementary material. 

% Lastly, we analyze the efficiency of different backbones, as detailed in Table~\ref{tab:back}. The results indicate that ResNet50 is the preferred backbone for the student model of MonoTAKD, offering higher AP, faster FPS, and lower FLOPs. Hence, we select ResNet50 as the backbone for experiments on the KITTI test, val and raw datasets. 

\vspace{-6pt}
\section{Discussion}
\label{sec:limitation}
To garner 3D perception from a 2D image, recent Mono3D methods can be categorized into three major strategies: (1) Extra training data. This method leverages LiDAR-generated auxiliary labels, such as depth maps~\cite{monodtr,monopgc,mononerd}, occupancy labels~\cite{occm3d}, temporal data~\cite{kin3d,stxd}, to improve the training process.
(2) Depth-guided. Using additional depth estimator~\cite{fusion1,fusion2,monoatt,monodetr} or pre-training on additional depth datasets~\cite{dd3d,dd3dv2} to improve depth estimation.
(3) Cross-modal. Leveraging LiDAR-based teacher to camera-based student~\cite{cmkd,unidis,2023disbev,labeldis}.
Fair comparisons are challenged by heterogeneous experimental settings and varying model architectures. The lack of standardized efficiency metrics (FLOPs, parameter counts) makes it difficult to assess the true cost-benefit and trade-offs. We advocate for future work to report such metrics to enable fair comparisons across architectures and backbones. 

\section{Conclusion}
\label{sec:conclusion}
In this paper, we propose a novel teaching assistant knowledge distillation for Mono3D (MonoTAKD) to improve the cross-modal distillation effectiveness.
Specifically, we observe that the camera-based student model struggles to learn 3D information due to a significant feature representation gap between LiDAR and camera modalities.
We alleviate this impediment by employing a camera-based teaching assistant model to deliver robust 3D visual knowledge through intra-modal distillation.
To enhance the 3D perception of the student model, we formulate the LiDAR-exclusive 3D spatial cues as residual features and then distill them to the student model through cross-modal residual distillation.
As a result, MonoTAKD establishes a new state-of-the-art benchmark on the KITTI 3D detection dataset. With its accurate yet cost-effective attributes, MonoTAKD poses a promising solution for autonomous driving applications.

\section{Acknowledgments}
This work is partially supported by the National Science and Technology Council, Taiwan under Grants NSTC-112-2221-E-A49-059-MY3 and NSTC-112-2221-E-A49-094-MY3.

% \newpage
{
    \small
    \bibliographystyle{ieeenat_fullname}
    \bibliography{main}
}

\clearpage
\renewcommand{\thesection}{\Alph{section}}
\renewcommand\thefigure{\Alph{section}\arabic{figure}}
\renewcommand\thetable{\Alph{section}\arabic{table}}
\setcounter{page}{1}
\setcounter{section}{0}
\setcounter{figure}{0}
\setcounter{table}{0}

\maketitlesupplementary

Due to the page constraint of the main paper, we provide more quantitative and qualitative results in this supplementary material, which is organized as follows:

\begin{itemize}
    \item Dataset description of the KITTI raw set in Section~\ref{sec:dataset_sup}.
    \item The implementation and training details for the KITTI3D and nuScenes datasets are documented in Section~\ref{sec:imp}.
    \item Justification and analysis of the TA model in Section~\ref{sec:mta}.
    \item More quantitative results for MonoTAKD in Section~\ref{sec:mqual}.
    \item More ablation studies for MonoTAKD in Section~\ref{sec:mabla}.
    \item Qualitative results for MonoTAKD in Section~\ref{sec:mquan}. 
\end{itemize}

\section{Datasets}
\label{sec:dataset_sup}
\noindent \textbf{KITTI Raw}.
The KITTI Raw dataset includes approximately 48K unlabeled data used for semi-supervised training. 
Following~\cite{cmkd,lpcg}, we train on the Eigen clean subset (22K) of the KITTI raw dataset and evaluate on the KITTI test set (3,769).
The evaluation metric and the implementation of KITTI raw are the same as the KITT3D dataset. 

\section{Implementation Details} 
\label{sec:imp}
For the KITTI3D dataset, we use a pre-trained Second~\cite{second} as the LiDAR-based teacher. Both the camera-based TA and camera-based student are derived from CaDDN~\cite{caddn}, using ResNet50 as their backbone.
In addition, we use PointPillar~\cite{pointpillar} as the BEV detector. 
Initially, we trained the TA model using a pre-trained model for 5 epochs.  
Then, a pre-trained teacher model and a frozen TA model are used to train the student model for another 60 epochs.
Training is performed with an NVIDIA Titan XP GPU in an end-to-end manner. We set the batch size to 2, and the learning rate is $2e^{-4}$ with the one-cycle learning rate strategy.
The IoU thresholds for the Car, Pedestrian, and Cyclist categories are 0.7, 0.5, and 0.5, respectively.
As for the discrete depth bins $D$, we set $D$ to 120, and the minimum and maximum depths are set to 2.0 and 46.8 meters, respectively. 

In the case of the nuScenes dataset, we adopt a pre-trained CenterPoint~\cite{centerpoint} as the LiDAR-based teacher and use BEVDepth~\cite{bevdepth} for both the TA and the student. 
Due to a higher resolution and a larger model size, we set the batch size to 8 and trained the models with eight NVIDIA V100 GPUs.
We set the learning rate of $2e^{-4}$ with a multi-step learning rate decay schedule and a decay rate of 0.1 and train the model for 25 epochs.

\section{Justification and Analysis of TA}
\label{sec:mta}
\noindent \textbf{Novelty of the TA.}
Unlike previous TAKD~\cite{takd}, relying on step-by-step distillation, we bypass this and simultaneously distill complementary knowledge to the student: 3D visual knowledge from the camera-based TA and precise LiDAR-exclusive 3D features from the LiDAR-based teacher. This approach presents a novel solution to the cross-modal distillation problem, which goes beyond addressing the differences in the model's architecture.
Experimental results show that MonoTAKD outperforms TAKD by 4.5\%, 3.8\%, and 2.7\% in $AP_{3D}$ for easy, moderate, and hard scenarios, as step-by-step distillation cannot bridge the modality gap and also complicates the training procedure.

\noindent \textbf{Quality and complexity of TA.} 
To ensure high-quality features from the TA model, we fine-tune it starting from a pre-trained camera-based model for 5 epochs, achieving rapid convergence within 3 hours (simple training procedure) due to the incorporation of the GT depth, as shown in Table~\ref{tab:ta_map}.
Additionally, since the TA model is excluded during inference, it does not affect the student's inference time.

\vspace{-6pt}
\begin{table}[h]
\centering
\footnotesize
\caption{Performance of our teacher model $\mathcal{T}$ and teaching assistant model $\mathcal{A}$. $\dagger$ represents the incorporation of the GT depth.}
\vspace{-6pt}
\begin{tabular}{c|c|c|ccc} 
\hline
\multirow{2}{*}{Model} & \multirow{ 2}{*}{Epochs} & \multirow{ 2}{*}{Training Time (hr)} & \multicolumn{3}{c}{$AP_{3D}$} \\ 
\cline{4-6}
                          &       &               & Easy   & Mod.   & Hard         \\ 
\hline
$\mathcal{T}$                   & N/A     &  pre-trained  & 87.68  & 76.32  & 73.28        \\ 
$\mathcal{A}$                        & N/A     &  pre-trained  & 23.47  & 16.31  & 13.84        \\ 
$\mathcal{A}^{\dagger}$                       & N/A     &  pre-trained  & 54.84  & 35.44  & 30.45        \\ 
$\mathcal{A}^{\dagger}$                       & 5     &  3  & 62.91  & 43.35  & 34.99        \\ 
$\mathcal{A}^{\dagger}$                       & 10    &  6  & 62.83  & 42.98  & 34.82        \\
\hline
\end{tabular}
\label{tab:ta_map}
\vspace{-12pt}
\end{table}

\noindent \textbf{Applicability of TA.}
One concern is whether depth maps are always available for training the TA model. Most autonomous driving datasets, including KITTI3D, nuScenes, and Waymo, provide 3D detection labels derived from LiDAR point clouds, which can be readily converted into GT depth maps for TA training. However, when depth maps are not directly accessible (e.g., radar 3D object detection), distance information can be used as an alternative. 

In summary, the overall performance, considering AP, training complexity, and model complexity, provides a superior solution compared to the existing Mono3D approach. Further discussion can be found in section~\ref{sec:limitation}.
% This effort is significantly smaller than the approaches highlighted in the discussion (see section~\ref{sec:limitation}). 

\section{More Quantitative Results}
\label{sec:mqual}

\begin{table*}[t]
\centering
\scriptsize
\caption{Experimental results for Pedestrian and Cyclist categories on the KITTI \textit{test} set. We use \textbf{bold} and \underline{underline} to indicate the best and the second-best results, respectively.} %Lastly, we highlight the performance gain over the second-best results in the last row.}
\resizebox{.98\linewidth}{!}{
\begin{tabular}{l|c|ccc|ccc}
\hline
\multirow{2}{*}{Method}    & \multirow{2}{*}{Venue}   & \multicolumn{3}{c|}{Pedestrian $AP_{3D}$/$AP_{BEV}$} & \multicolumn{3}{c}{Cyclist $AP_{3D}$/$AP_{BEV}$} \\
               & & Easy           & Mod.          & Hard          & Easy           & Mod.         & Hard        \\
\hline                           
% MonoPSR~\cite{monopsr}   & CVPR 2019 & 6.12/7.24    & 4.00/4.56   & 3.30/4.11   & 8.37/9.87    & 4.74/5.78  & 3.68/4.57 \\
MonoATT~\cite{monoatt} & CVPR 2023  & 10.55/11.63  & 6.66/7.40   & 5.43/6.56   & 5.74/6.73    & 3.68/4.44  & 2.94/3.75 \\
Cube R-CNN~\cite{omni-cube} & CVPR 2023  & 11.17/11.67  & 6.95/7.65   & 5.87/6.60   & 3.65/5.01    & 2.67/3.35  & 2.28/3.32 \\
CaDDN~\cite{caddn} & CVPR 2021 & 12.87/14.72  & 8.14/9.41   & 6.76/8.17   & 7.00/9.67    & 3.41/5.38  & 3.30/4.75 \\
DD3D~\cite{dd3d} & ICCV 2021  & 13.91/15.90  & 9.30/10.85  & 8.05/8.05   
                & 2.39/3.20    & 1.52/1.99   & 1.31/2.39 \\
% CMKD~\cite{cmkd}  & \underline{17.79}/\underline{20.42}  & 11.69/13.47  & 10.09/\underline{11.64}   
%                 & \underline{9.60}/\underline{12.53}  & \underline{5.24}/\underline{7.24}  & \underline{4.50}/\underline{6.21} \\
% CMKD~\cite{cmkd}  & \underline{13.94}/\underline{16.03}  & 8.79/10.28  & 7.42/\underline{8.85}   
%                 & \underline{12.52}/\underline{14.66}  & \underline{6.67}/\underline{8.15}  & \underline{6.34}/\underline{7.23} \\
MonoNerd~\cite{mononerd}  & ICCV 2023 & 13.20/15.27  & 8.26/9.66  & 7.02/8.28   
                & 4.79/5.24    & 2.48/2.80   & 2.16/2.55 \\
MonoUNI~\cite{monouni} & NeurIPS 2023  & \underline{15.78}/\underline{16.54}  & \underline{10.34}/\underline{10.90}  & \underline{8.74}/\underline{9.17}   
                & 7.34/8.25  & \underline{4.28}/\underline{5.03}  & \underline{3.78}/\underline{4.50} \\

OccupancyM3D~\cite{occm3d} & CVPR 2024 & 14.68/\underline{16.54}  & 9.15/10.65  & 7.80/9.16   
                &  \underline{7.37}/\underline{8.58}  & 3.56/4.35 & 2.84/3.55 \\

\hline
\textbf{MonoTAKD} & - & \textbf{16.15}/\textbf{19.79}  & \textbf{10.41}/\textbf{13.62} & \textbf{9.68}/\textbf{11.92}
                & \textbf{13.54}/\textbf{16.90} & \textbf{7.23}/\textbf{9.42}  & \textbf{6.86}/\textbf{8.29} \\
% \textit{Improvement} & \textit{+2.21}/\textit{+3.76}  & \textit{+1.11}/\textit{+2.77}   & \textit{+1.63}/\textit{+3.07}  & \textit{+1.02}/\textit{+2.24}    & \textit{+0.56}/\textit{+1.27}  & \textit{+0.52}/\textit{+1.06} \\

\hline
\end{tabular}
}
\label{tab:ped_cyc}
\end{table*}

\begin{table*}[t]
\centering
\scriptsize
\caption{Experimental results on the KITTI \textit{test} set for the Car category, leveraging unlabeled data. We use {\textbf{bold}} and \underline{underline} to indicate the best and the second-best results, respectively.}
\resizebox{0.9\linewidth}{!}{
\begin{tabular}{l|c|c|ccc|ccc}
\hline
\multirow{2}{*}{Method}  & \multirow{2}{*}{Venue} & \multirow{2}{*}{Extra Data} & \multicolumn{3}{c|}{$AP_{3D}$} &  \multicolumn{3}{c}{$AP_{BEV}$}     \\
% \cline{4-6} \cline{7-9}
                         &         &            & Easy  & Mod.  & Hard  & Easy   & Mod.  & Hard  \\
\hline
% \hline \\[-2ex]
% \multicolumn{9}{c}{Semi-Supervised Learning} \\
LPCG~\cite{lpcg}          & ECCV 22 & \multirow{3}{*}{Raw} & 25.56 & 17.80 & 15.38 & 35.96  & 24.81 & 21.86 \\
Mix-Teaching~\cite{mixtea}& CSVT 23 &            & 26.89 & 18.54 & 15.79 & 35.74  & 24.23 & 20.80 \\
CMKD~\cite{cmkd}          & ECCV 22 &            & \underline{28.55} & \underline{18.69} & \underline{16.77} & \underline{38.98} & \underline{25.82} & \underline{22.80} \\
\hline
\textbf{MonoTAKD}  & -  &  Raw  & \textbf{29.86} & \textbf{21.26} & \textbf{18.27} & \textbf{43.83} & \textbf{32.31} & \textbf{28.48} \\
% \textbf{MonoLTKD (Ours)}  & -       &  Raw       & \textbf{31.28} & \textbf{21.83} & \textbf{18.87} & \textbf{45.37} & \textbf{33.30} & \textbf{33.39} \\
\hline
\end{tabular}
}
\label{tab:semi}
% \vspace{-4pt}
\end{table*}

% 29.75 19.09 16.93 39.41 26.02 22.76

\noindent \textbf{Results for Pedestrian and Cyclist.}
We present a detailed comparison with other state-of-the-art methods for the non-car categories on the KITTI test set.
Table~\ref{tab:ped_cyc} demonstrates that MonoTAKD outperforms other methods not only in the Car category but also in the Pedestrian and Cyclist categories.
This success indicates that the approach is well-suited for a broad range of autonomous driving applications, including tasks like trajectory prediction.

\noindent \textbf{Results on KITTI raw.}
To improve the transferability and to generalize the application of MonoTAKD on real-world scenes, we explore the performance of MonoTAKD in a semi-supervised manner.
% As illustrated in Table~\ref{tab:semi}, our MonoTAKD surpasses CMKD by 1.31/3.85, 2.57/6.49, and 1.50/5.68 in $AP_{3D}$/$AP_{BEV}$ across all three difficulty levels, respectively.
As illustrated in Table~\ref{tab:semi}, our MonoTAKD outperforms CMKD in $AP_{3D}$/$AP_{BEV}$ across all three difficulty levels, respectively.

Owing to MonoTAKD's outstanding performance in semi-supervised settings, it is evident that our distillation method adeptly extracts valuable 3D features from unlabeled data. Thus, MonoTAKD can provide comprehensive guidance for the student model across all difficulty levels.

\section{More Ablation Studies}
\label{sec:mabla}

\noindent \textbf{Backbone choices on KITTI3D.} 
We analyze the backbone choice of our MonoTAKD in Table~\ref{tab:back}.
According to the table, Swin-T, a transformer-based backbone, exhibits higher FLOPs and underperforms in both speed and accuracy. We believe the performance drop is because of the heterogeneity of the architecture between teacher and student (CNN and Transformer). Conversely, MobileNetV3, a lightweight backbone, excels in speed and efficiency with lower FLOPs but has a trade-off with lower accuracy.

After comparing ResNet50 and ResNet101, we determined that ResNet50 is the optimal backbone for the student model, delivering enhanced performance with higher AP, improved FPS, and reduced FLOPs. This finding highlights that in Mono3D tasks, a larger or more complex backbone does not necessarily translate to better performance.
Note that, we only compare the FLOPs of the backbone. The total FLOPs can be found in Table 6.

\begin{table}[ht]
\centering
\footnotesize
\caption{Comparison of MonoTAKD with different backbones.}
% \vspace{-5pt}
\resizebox{\linewidth}{!}{
\begin{tabular}{l|c|c|ccc} 
\hline
\multirow{2}{*}{Backbone} & \multirow{2}{*}{Speed (FPS)} & \multirow{2}{*}{FLOPs (G)} & \multicolumn{3}{c}{$AP_{3D}$} \\ 
\cline{4-6}
& & & Easy & Mod. & Hard \\ 
\hline
Swin-T        & 5.8  & 16.7 & 31.57 & 19.33 & 17.65 \\
MobileNetV3   & 13.8  & 3.4 & 26.11 & 16.87 & 13.92 \\
ResNet101     & 9.2   & 4.3  & 33.07 & 21.54 & 19.16 \\
\textbf{ResNet50} & \textbf{11.9}  & \textbf{4.1} & \textbf{34.36} & \textbf{22.61} & \textbf{19.88} \\

\hline
\end{tabular}
}
\label{tab:back}
\end{table}
% \vspace{-10pt}
% \vspace{-5pt}

\section{Qualitative Results}
\label{sec:mquan}
\begin{figure*}[t]
\centering
    % Second row with three subfigures
    \begin{subfigure}{0.3\textwidth}
    \includegraphics[width=\textwidth]{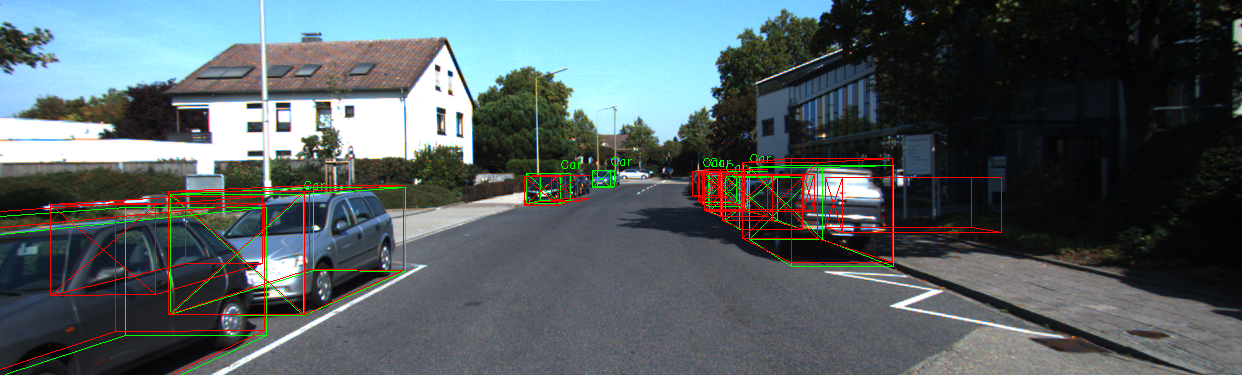}
    \label{fig:sub4}
    \end{subfigure}
    \hfill
    \begin{subfigure}{0.3\textwidth}
    \includegraphics[width=\textwidth]{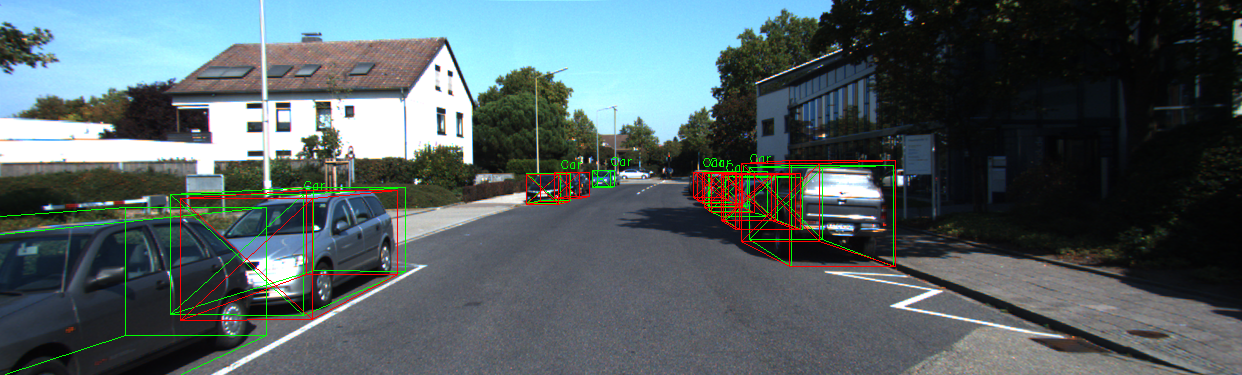}
    \label{fig:sub5}
    \end{subfigure}
    \hfill
    \begin{subfigure}{0.3\textwidth}
    \includegraphics[width=\textwidth]{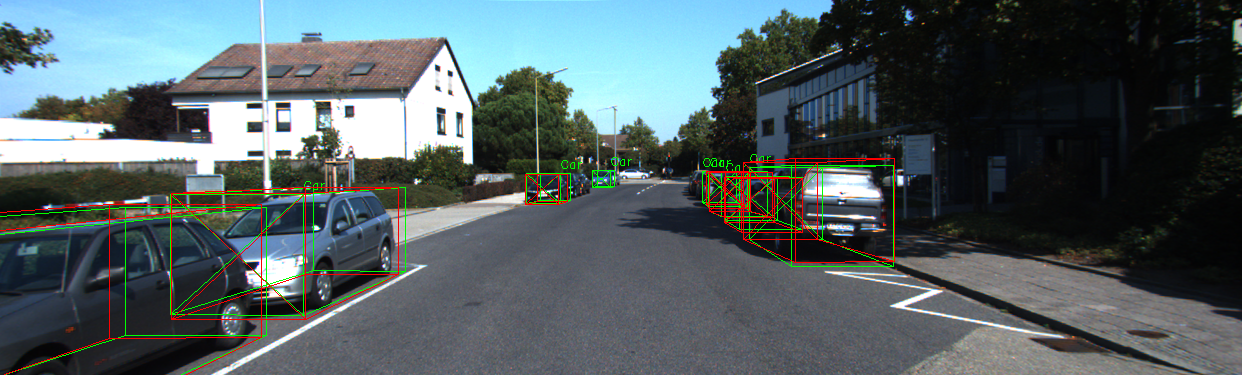}
    \label{fig:sub6}
    \end{subfigure}
    % First row with three subfigures
    \begin{subfigure}{0.3\textwidth}
    \includegraphics[width=\textwidth]{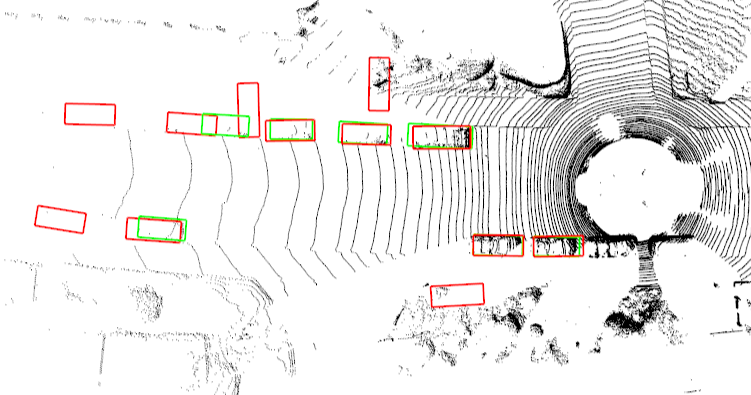}
    \label{fig:sub1}
    \caption{CMKD}
    \end{subfigure}
    \hfill
    \begin{subfigure}{0.3\textwidth}
    \includegraphics[width=\textwidth]{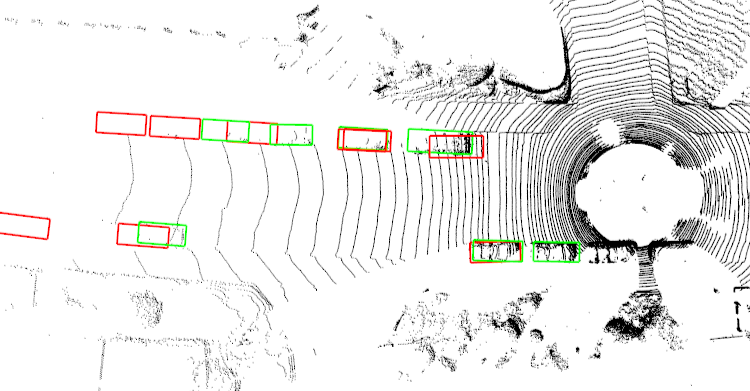}
    \label{fig:sub2}
    \caption{MonoDETR}
    \end{subfigure}
    \hfill
    \begin{subfigure}{0.3\textwidth}
    \includegraphics[width=\textwidth]{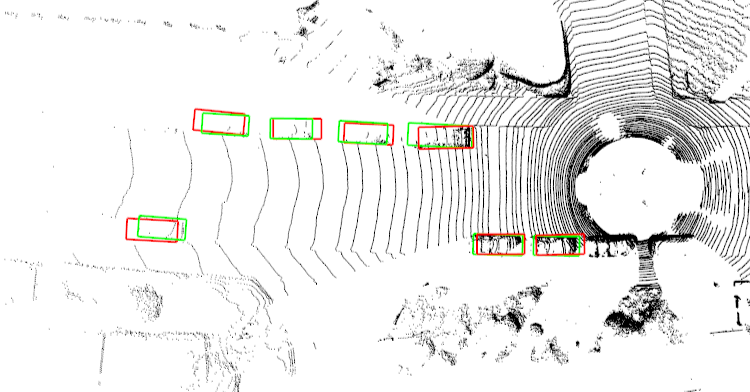}
    \label{fig:sub3}
    \caption{MonoTAKD (ours)}
    \end{subfigure}

    \caption{Qualitative results on KITTI \textit{val} set for the Car category. We compare the qualitative results among CMKD~\cite{cmkd}, MonoDETR~\cite{monodetr}, and our proposed MonoTAKD. The first and second rows represent detection results from a camera frontal view and a BEV, respectively. We use green and red boxes to indicate the ground truth and prediction bounding boxes.}
    \label{fig:3d_bev}  
\end{figure*}

We compare our visualization results with state-of-the-art methods, CMKD~\cite{cmkd} and MonoDETR~\cite{monodetr}, for both 3D object and BEV detection in Fig.~\ref{fig:3d_bev}.
MonoTAKD comparatively has the best-fitted bounding box size estimation and the most accurate 3D localization among the three methods. 

\begin{figure*}[ht]
\centering
            \includegraphics[width=.95\textwidth]{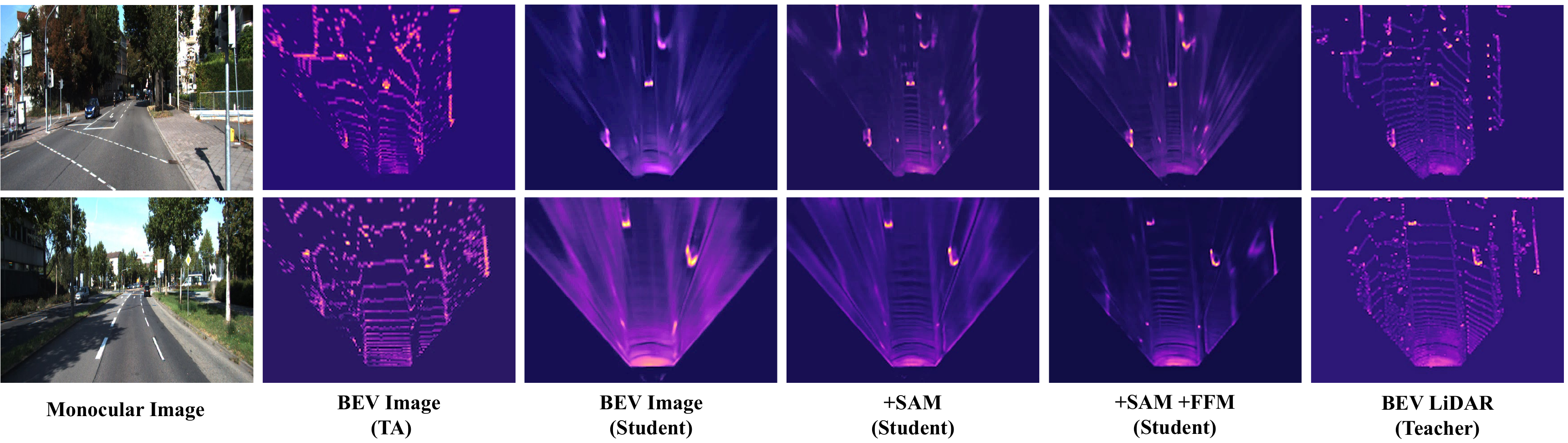}
            \caption{Visualization of the BEV features from the teacher, TA, and the two distillation branches of the student model on KITTI \textit{val} set.}
            \label{fig:vis_bev}  
\end{figure*}

Lastly, Fig.~\ref{fig:vis_bev} presents the BEV features of the teacher, TA, and the student.
Notably, the student's BEV image exhibits distortion and blurriness. However, with the help of SAM and FFM modules, the student's BEV features successfully align more closely to resemble the BEV LiDAR features. This visual comparison illustrates how the proposed approaches collectively contribute to improving the student's 3D perception.

\end{document}